# A multi-model-based deep learning framework for short text multiclass classification with the imbalanced and extremely small data set

Jiajun Tong
China University of Mining and Technology,School of Computer Science and Technology
E-mail: tb20170008b4@cumt.edu.cn

Zhixiao Wang
China University of Mining and Technology,School of Computer Science and Technology
E-mail: zhxwang@cumt.edu.cn

Xiaobin Rui
China University of Mining and Technology,School of Computer Science and Technology
E-mail: ruixiaobin@cumt.edu.cn

## Abstract:

Text classification plays an important role in many practical applications. In the real world, there are extremely small datasets. Most existing methods adopt pre-trained neural network models to handle this kind of dataset. However, these methods are either difficult to deploy on mobile devices because of their large output size or cannot fully extract the deep semantic information between phrases and clauses. This paper proposes a multimodel-based deep learning framework for short-text multiclass classification with an imbalanced and extremely small data set. Our framework mainly includes five layers: the encoder layer, the word-level LSTM network layer, the sentence-level LSTM network layer, the max-pooling layer, and the softmax layer. The encoder layer uses DISTILBERT to obtain context-sensitive dynamic word vectors that are difficult to represent in traditional feature engineering methods. Since the transformer part of this layer is distilled, our framework is compressed. Then, we use the next two layers to extract deep semantic information. The output of the encoder layer is sent to a bidirectional LSTM network, and the feature matrix is extracted hierarchically through the LSTM at the word and sentence level to obtain the fine-grained semantic representation. After that, the max-pooling layer converts the feature matrix into a lower-dimensional matrix, preserving only the obvious features. Finally, the feature matrix is taken as the input of a fully connected softmax layer, which contains a function that can convert the predicted linear vector into the output value as the probability of the text in each classification. Extensive experiments on two public benchmarks demonstrate the effectiveness of our proposed approach on an extremely small data set. It retains the state-of-the-art baseline performance in terms of precision, recall, accuracy, and F1 score, and through the model size, training time, and convergence epoch, we can conclude that our method can be deployed faster and lighter on mobile devices.

# 1. Introduction

Text classification plays an important role in many practical applications [1]. Especially with recent breakthroughs in natural language processing (NLP) and text mining, it is widely used in many information processing systems, such as search engines [2][3] and question answering systems [4][6]. Generally, text classification can be divided into 3 steps: text preprocessing, feature extraction, and text representation, as shown in **Figure 1**.

In the real world, many datasets are unbalanced and extremely small, which is difficult to solve with traditional machine learning methods [6]. For example, questions about fires or accidents are usually less common than questions about consumer disputes about property services, but these data are too important to be ignored because of their scarcity and implications. How to efficiently and accurately classify those texts into specific categories according to the content of the corpus and improve the user experience has become an urgent problem to be solved.

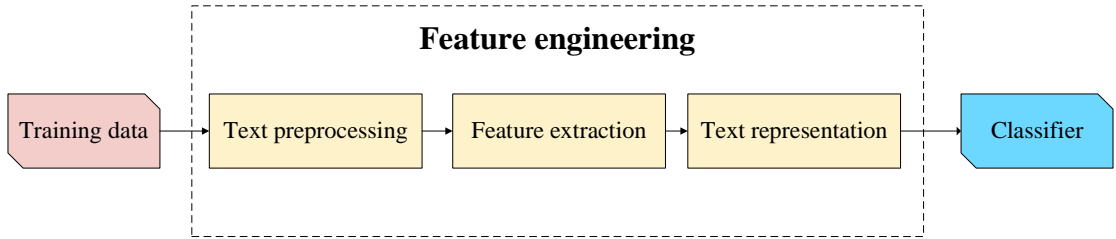

**Figure 1 Overview of a text classification pipeline**

To solve the above problem, most existing methods are based on the pre-trained neural network, which can utilize a large amount of unlabelled data to learn common language representations and mine the relationship between context and target aspects [7]. Some of the most prominent methods are based on BERT [8]. Song Y [9] explored the potential of utilizing BERT intermediate layers to enhance the performance of the fine-tuning part. Wan C X [10] proposed a combination model named BERT-CNN for the classification of candidate causal sentences. The number of encoder layers in BERT leads to not only a large number of training parameters but also a large output model size. Therefore, some distillation methods [11] have been proposed. Xiong G [12] proposed a sentiment analysis model based on distillation bidirectional encoder representations from transformers combined with multiscale convolution. Adel H [13] presented an alternative event detection model based on DistilBERT and the Hunger Games search algorithm. The core idea of distillation is called teacher-student learning, where the student can reproduce the large model, and the teacher learns dark knowledge.

However, these methods are either difficult to deploy to mobile devices because of their output model size and require professional computing resources such as GPUs, or cannot fully extract the deep semantic information between phrases and clauses, leading to a low recall rate and accuracy.

To reduce the size of the model and maintain the classification accuracy of the model. This article proposes a multimodel-based deep learning framework for short text multiclass classification with an imbalanced and extremely small dataset. Our framework consists mainly of four layers. First, we use the pre-trained DistilBERT as the encoder layer to obtain the context-sensitive dynamic word embeddings, including the token embeddings, position embeddings, and segment embeddings, as the input of the bidirectional LSTM network. Second, hidden features between phrases and clauses in the text are extracted through word-level and sentence-level LSTM networks to obtain the deep semantic information of sentences while stored as a feature matrix. To reduce the parameters and network complexity, we add a max-pooling layer to convert the feature matrix into a lower-dimensional matrix. Finally, for the multiclass classification task, the softmax layer normalizes multiple values obtained by the neural network to make the values between [0,1] so that the results can be explained.

The main contributions of our paper can be summarized as follows:

1. We built a multimodel-based framework combined with DistilBERT and BI-LSTM for the task of short text multiclass classification. Our method ensures accuracy and compresses the size of the output, which can be better applied to various mobile devices, such as smart appliances or smart cars.
2. In the encoder layer, DistilBERT can convert the input to dynamic word embeddings, including token embeddings, segment embeddings, and position embeddings, which can solve the polysemous word problem while reducing the training parameters and the output size.
3. The BI-LSTM is used as the feature extraction layer to extract the implicit features as fine-grained as possible through the word-level and sentence-level LSTM network to enhance the classification performance.

## 2. Related work

As a fundamental topic in the NLP field, there is much research related to text classification. On the one hand, text categorization tasks are usually based on large amounts of annotated data, no matter whether they are supervised learning or self-supervised learning methods. Ghiassi M [14] presented an integrated solution that combines a new clustering algorithm, However, another clustering algorithm (YAC2), with a domain transferable feature engineering approach for Twitter sentiment analysis and spam filtering of YouTube comments. Kim Y [15] proposes a question-answer method to automatically provide users with infrastructure damage information from textual data. STITINI O [16] concludes that the linkage between contextual information and classification enhances and improves the recommendation results. Yang DU [17] proposed a new automated defect text classification system (AutoDefect) based on a convolutional neural network (CNN) and natural language processing (NLP) using hierarchical two-stage encoders. Ma Y [18] presented a level-by-level HMTC approach based on the bidirectional gated recurrent unit network model together with hybrid embedding used to learn the representation of the text level-by-level. These methods explore the method of applying text classification to practical tasks.

However, many datasets are imbalanced and extremely small in the real world. Bilal S F [19] proposed a churn prediction model based on a combination of clustering and classification algorithms using an ensemble. Wang K [20] proposes TkTC: a framework for top k text classification, where a novel loss function simultaneously considers the position of the ground truth label and the number of predictions. Zhu Y, [21] proposes a simple short-text classification approach that makes use of prompt learning based on knowledgeable expansion, which can consider both the short text itself and the class name while expanding the label word space.

Although those short text classification methods based on the pretraining model can show good performance in the short text classification task, these models would generate a large output, which is unsuitable for mobile devices. Therefore, solutions based on DistilBERT are inspired. Andersen J S [22] proposed a framework for the efficient, in-operation moderation of classifier output to maximize the accuracy and increase the overall acceptance of text classifiers. Chang, J W [23] developed a universal financial fraud awareness model to avoid these cases escalating to the level of crime. However, the above methods cannot guarantee the application performance on mobile devices, in which the memory space and computation resources are often limited.

In this article, we propose a multi-model-based framework combining the advantages of DistilBERT and BI-LSTM to reduce the model size in small dataset tasks while maintaining the performance of classification.

# 3 Distilbert BI-LSTM predictor

In this section, we propose a short-text classification framework based on multi-model learning combined with distillation encoder representations with bidirectional LSTM, which trains tasks in a deep learning model, and the deep learning model combines neural networks with different structures to benefit from it. First, this model inputs texts into the encoder layer to obtain segment embeddings, position embeddings, and token embeddings. Second, the batched embeddings are sent into the forward LSTM to extract the word-level features, and the backward LSTM to extract the sentence-level features. Since the BI-LSTM extracts high-dimensional features, we adopt the max pooling operation to sharpen the feature matrix and finally output the classification results through the softmax layer. The overall structure of the model is shown in **Figure 2**, and the rest of this section will introduce this method in detail.

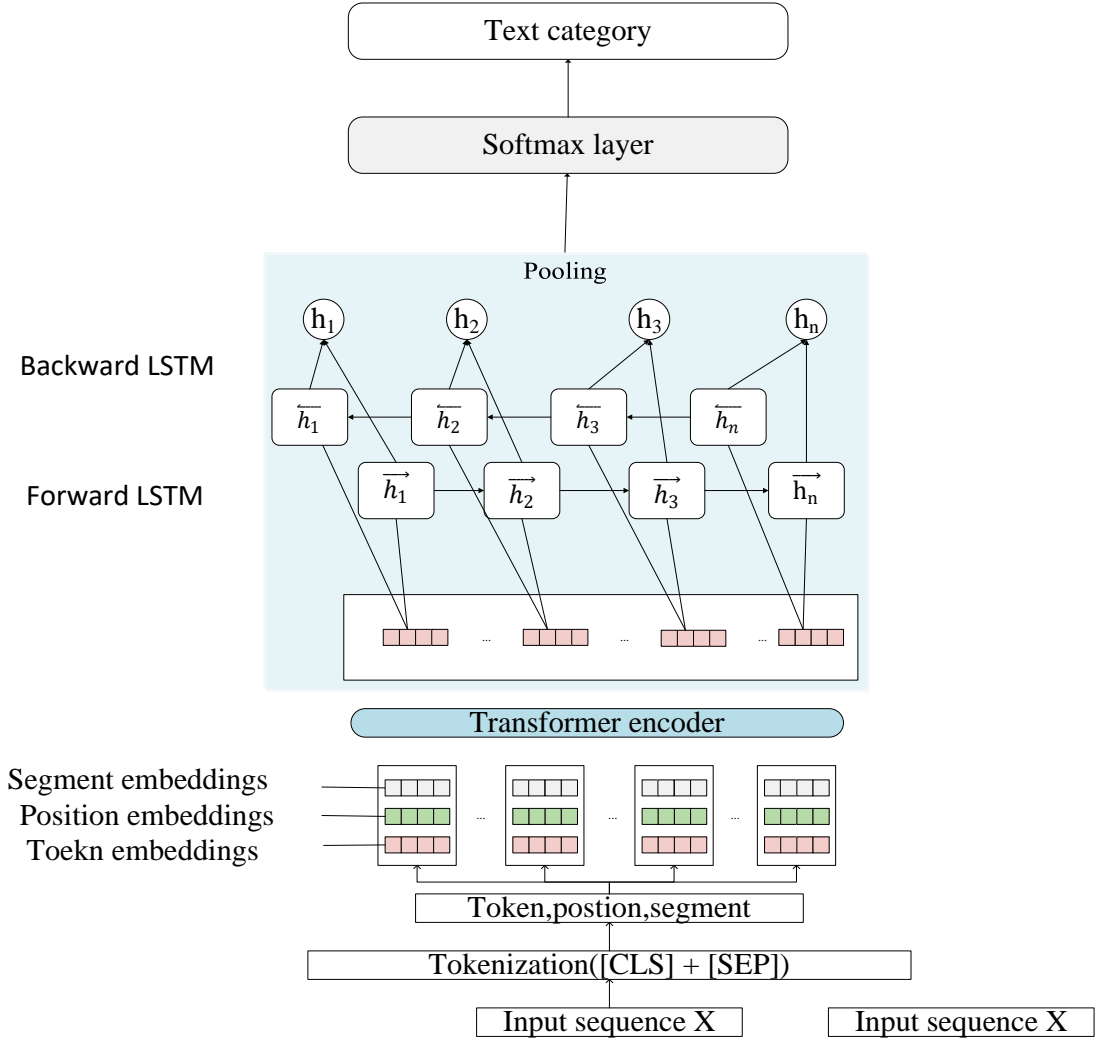

**Figure 2 The framework based on multimodel-based deep learning**

## 3.1 Encoder layer

We remove the first input layer and have 6 encoder layers of DistilBERT[11]. Dilbert uses knowledge distillation to minimize the BERT base model parameters by 40%, making the inference 60% faster. Thus, the number of transformer layers (encoders) in the BERT base (12 layers) has been reduced to six. The first token (CLS) vector of each encoding layer can be used as a sentence vector. We can abstractly understand that the shallower the encode layer, the better the sentence vector. It represents low-level semantic information, and the deeper it is, it represents high-level semantic information. We aim to obtain semantic features while preserving relevant word features. The specific method of the model is to use the CLS vectors of layers 1 to 6 as input of the LSTM for classification.

Semantic representation is assumed to be the contextual embedding learned by the token [CLS], which serves as the semantic representation of the input tweet X. The task is formulated as a multiclass classification problem. Therefore, the probability of X being classified as class c (i.e., an event) is predicted as the Softmax function used in Equation (1).

$$Pr(c|X) = Softamax(WT \cdot X) \qquad (1)$$

where W is the weight matrix learned during the fine-tuning of the pre-trained model used during the initialization of the feature extractor model. r is the number of classes. It is worth noting that the first five transformer layers in the pre-trained model are not trainable. We only fine-tuned the last transformer layer (encoder) of the pre-trained model and replaced the classification layer with two fully connected layers for feature extraction and classification.

## 3.2 Word level LSTM network layer

In the short text classification task, each input unit may have different degrees of impact on the final classification results. The LSTM network can filter the information from the input units. Based on RNN, the network adds gates for selecting information, including an input gate, output gate, forgetting gate, and a memory unit to store information and update information through different gates.

To capture as much fine-grained classification information as possible in a small amount of text, this paper extracts and represents the semantic features of sentences through word- and sentence-level LSTM networks. The updating process of the word-level LSTM network is as follows: in time step T, the forgetting gate FWT will first judge which information can be forgotten according to the input XT of the current time and the output HWT-1 of the previous time, for example, the root units "##ing", "##ed" and crown words segmented. This can be described as follows:

$$f_t^w = \sigma(W_{xf}x_t + W_{hf}h_{t-1}^w + b_f) \qquad (2)$$

where $W_*$ represents the weight matrix and $b_*$ presents the bias term, all of which are network parameters to be learned; $\sigma$ Activate the sigmoid function.

Then, the input gate $i_t^w$ will determine which information has a great impact on the classification label and needs to be updated. For example, some words have a great impact on the classification, such as "result", "case" and "great". The network will create a new memory cell candidate value $\tilde{C}_t^w$ through the function. It can be described as follows:

$$i_t^w = \sigma(W_{xi}x_t + W_{hi}h_{t-1}^w + b_i) \qquad (3)$$
$$\tilde{C}_t^w = tanh(W_{xc}x_t + W_{hc}h_{t-1}^w + b_c) \qquad (4)$$

en, when updating the memory cell $C_t^w$, it is necessary to combine the forget gate $f_t^w$ and the memory cell $C_{t-1}^w$ at the last time for information screening. The information with the quantity product $f_t^w \cdot C_{t-1}^w$ close to 0 will be discarded and added with $i_t^w \cdot \tilde{C}_t^w$ to obtain the updated memory cell $C_t^w$ to store the latest information. $C_t^w$ can be described as:

$$C_t^w = f_t^w \cdot C_{t-1}^w + i_t^w \cdot \tilde{C}_t^w \qquad (5)$$

Finally, the output gate $o_i^w$ combines the memory unit $C_t^w$ and the last time output $h_{t-1}^w$ with the current time input $x_t$ to calculate the word-level time step t as the output $h_t^w$. The calculation method is:

$$o_i^w = \sigma(W_{xo}x_t \cdot W_{ho}h_{t-1}^w + b_o) \qquad (6)$$
$$i_t^w = o_i^w \times \tanh(C_t^w) \qquad (7)$$

### 3.3 Sentence level LSTM network layer

For the feature information output from the pre-trained model, we use another LSTM to extract its fine-grained semantics and representation process

For the clause $S = \{s_1, ..., s_q, ..., s_n\}$ in sentence $s_q = \{w_{q1}, ..., w_{qp}, ..., w_{qm}\}$, the implicit feature $h_{sq}^w$ obtained by the word-level LSTM network and the lexical feature $B_{sq}$ obtained by BERT are spliced as the semantic feature representation $h_{sq}^w$ of the clause, that is,

$$h_{sq}^s = [(1-\lambda)B_{sq}, \lambda h_{sq}^w] \quad q \in [1,n] \tag{8}$$
$$h_{sq}^w = LSTM(w_{qp}) \quad q \in [1,m] \tag{9}$$

where $\lambda$ The parameter represents the weight. Taking sentence $S = \{s_1, ..., s_q, ..., s_n\}$ as a sequence composed of N clauses, its feature vector is expressed as:

$$S = \begin{cases} h_{s1}^w & n=1 \\ [h_{s1}^s, h_{s2}^s, ..., h_{sn}^s] & n \geq 2 \end{cases} \tag{10}$$

The feature vector is updated through the sentence-level LSTM network to obtain the long sentence semantic representation HT, and its process is similar to the word-level LSTM network, that is:

$$h_t = LSTM(S(h_{st}^s)) \quad t \in [1,n] \tag{11}$$

### 3.4 Max pooling layer

To improve the sharpening effect of pooling, we adopt the max-pooling operation. Because the BI-LSTM will eventually be connected to the fully connected layer and the number of neurons needs to be determined in advance, if the input length is uncertain, it is difficult to design its network structure. We use max-pooling to process the input X of uncertain length into a fixed-length input. Each filter takes only one value through the pooling operation. The number of neurons in the pooling layer corresponds to the number of filters so that the number of neurons in the eigenvector can be fixed. After the pooling operation, 2D or 1D arrays are usually converted to a single value, which can also reduce the number of parameters of a single filter or the number of hidden layer neurons for later use in the convolution layer or fully connected hidden layer.

### 3.5 Softmax layer

The output of the sentence-level LSTM network is transmitted to the full connection layer as the last semantic feature of the sentence, and the sentence classification result is obtained through the normalization operation of the Softmax function. $p(l_i|h)$ presents the probability of text $S$ in the $i$th classification. Its calculation method is as follows:

$$p(l_i|h) = Softmax(h_i)_i = \frac{\exp(w_i h_t + b_i)}{\sum_{j=1}^m \exp(w_j h_t + b_j)} \tag{13}$$

where $w_*$ and $b_*$ represent the weight matrix and offset term, respectively, and m represents the total number of tags. In this paper, the gradient descent algorithm is used to optimize the model and the cross entropy function is used to calculate the model loss and update the model parameters. The calculation method of loss function L is:

$$L = -\frac{1}{m}\sum_{i=1}^{m} y_i \log(p(l_i|h)) + \varphi\|w_*\|^2 \tag{14}$$

where $y_i$ represents the value of the digital vector of the real label of the text in the $i$ dimension and $\varphi$ represents the L2 regularization parameter.

# 4 Experiment

## 4.1 Dataset and metrics

We use three different datasets to evaluate our methods. The HUFF datasets contain approximately 200k news headlines from 2012 to 2018 obtained from HuffPost. This data set could be used to identify tags for untracked news articles or to identify the type of language used in different news articles. The COVID-Q [21] dataset has 1,690 questions about COVID, which are labeled into 16 unique categories. Considering that many text classification datasets need at least tens of thousands, this is a very small dataset. Finally, we evaluated it on a private dataset to see how it performed on a real-world application.

The notions of precision, recall, and F measures can be applied to each label independently in the multiclass task. Especially for an imbalanced classification problem with more than two classes, precision is calculated as the sum of true positives across all classes divided by the sum of true positives and false positives across all classes. where $y$ represents the set of $predicted(sample, label)$ pairs; $\hat{y}$ represents the set of $true(sample, label)$ pairs; $L$ the set of label pairs; $S$ represents the set of label pairs; and $y_s$ represents the subset of $y$ with sample $s$, i.e., $y_s := \{(s', l) \in y | s' = s\}$; $y_l$ represents the subset of $y$ with label $l$; Similarly, $\hat{y}_s$ and $\hat{y}_l$ are subsets of $\hat{y}$; $P(A, B := \frac{|A \cap B|}{|A|})$ for some sets $A$ and $B$; $R(A, B := \frac{|A \cap B|}{|A|})$ (Conventions vary on handling $B = \emptyset$; this implementation uses $R(A, B) := 0$, and similar for $P$). The F measurements can be defined as follows:

$$F_\beta(A, B) := (1 + \beta^2) \frac{P(A, B) \times R(A, B)}{\beta^2 P(A, B) + R(A, B)} \tag{15}$$

To give different weights to each type, we use weighted precision, recall and f1-score. **Precision** is calculated as the sum of true positives across all classes divided by the sum of true positives and false positives across all classes. Then, we define precision as follows:

$$\frac{1}{\sum_{l \in L} |\widehat{yl}|} \sum_{l \in L} |\widehat{yl}| P(yl, \widehat{yl}) \tag{16}$$

For example, we may have an imbalanced multiclass classification problem where the majority class is the negative class, but there are two positive minority classes: class 1 and class 2. Precision can quantify the ratio of correct predictions in both positive classes. **The recall** is calculated as the sum of true positives in all classes divided by the sum of true positives and false negatives across all classes. The recall is calculated as follows:

$$\frac{1}{\sum_{l\in L}|\widehat{yl}|}\sum_{l\in L}|\widehat{yl}|R(yl,\widehat{yl}) \tag{17}$$

We did not calculate an overall F1 score. Instead, we calculate the **F1 score** per class in a one-vs-rest manner. In this approach, we rate the success of each class separately, as if there are distinct classifiers for each class. The F1 score is defined as

$$\frac{1}{\sum_{l\in L}|\widehat{yl}|}\sum_{l\in L}|\widehat{yl}|F_\beta(yl,\widehat{yl}) \tag{18}$$

## 4.2 Results and Discussion

According to the method in Section 3, we construct a model and train the model on three different datasets. We verified the ensemble learning and multimodel methods on two public datasets. Moreover, to verify the performance of our model on small-scale datasets without changing the distribution of data, since there are approximately 200k lines of data in HUFF, we randomly extract 5% of the data from each type of data. At the same time, we deliberately keep the long tail problem for the data; that is, we do not limit the number of samples from each classification. Extract data naturally. We segment the data set according to 64% of the training set, 16% of the verification set, and 20% of the test set.

### 4.2.1 Comparison with ensemble modeling

Ensemble learning achieves better prediction performance by training multiple classifiers and combining these classifiers. The result of ensemble learning is usually better than a single model, which can reduce overfitting while improving accuracy, and the greater the difference between models (diversity), the more significant the improvement effect. This difference can be reflected in data, features, models, parameters, etc.

Encouraged by the good performance of the SVM model in the short-text classification task, we built an ensemble model to enhance this method, which compared 10 popular classifiers to evaluate the mean accuracy of each of them by a stratified k-fold cross-validation procedure. Thus, for the fine-tuning part, we performed a grid search optimization for 5 classifiers. Finally, we choose a voting classifier to combine the predictions, as **Figure 3** shows.

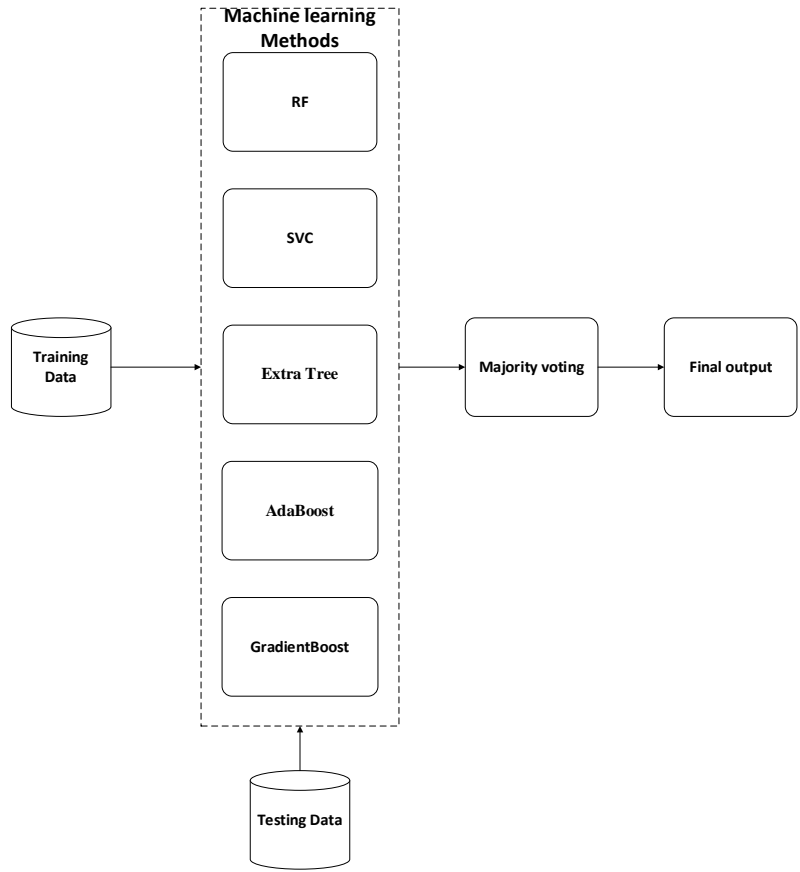

**Figure 3 Five classifiers combined with the Ensemble model**

Considering that the correlation between base models should be as small as possible, the performance gap should not be too large. We chose the SVC, AdaBoost, RandomForest, ExtraTrees, and GradientBoosting classifiers for ensemble modeling. We used learning curves to see the effect of overfitting the training set and the effect of the training size on the accuracy shown in **Figure 4**.

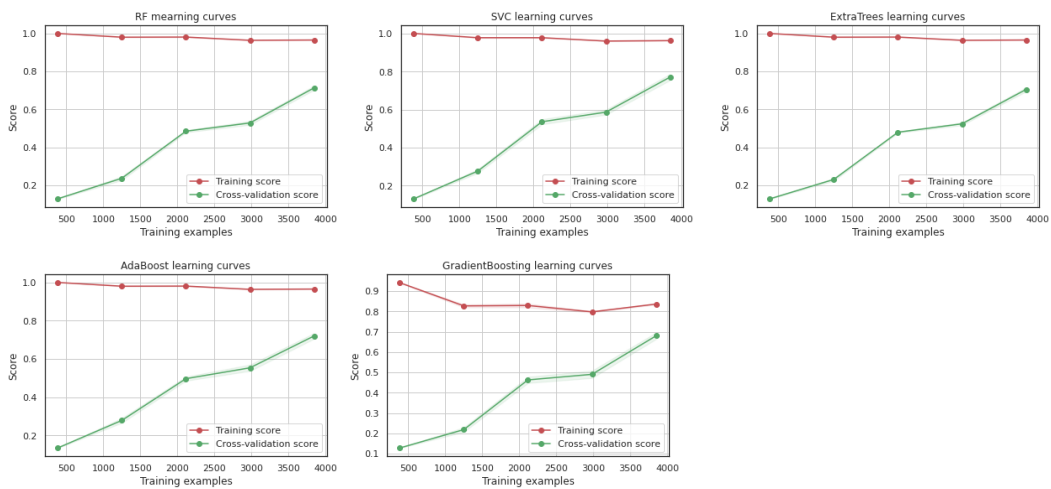

**Figure 4 Learning curves of each method**

GradientBoosting and Adaboost classifiers tend to overfit the training set. According to the growing cross-validation curves, GradientBoosting and Adaboost could perform better with more training examples. The SVC and ExtraTrees classifiers seem to better generalize the prediction since the training and cross-validation curves are close together. By comparing the ensemble model with our Distilbert BI-LSTM predictor, which is known as DBLP, on the HUFF dataset, the confusion matrix results are shown in **Figure 5**. DBLP improved in almost all categories and achieved a higher hit rate.

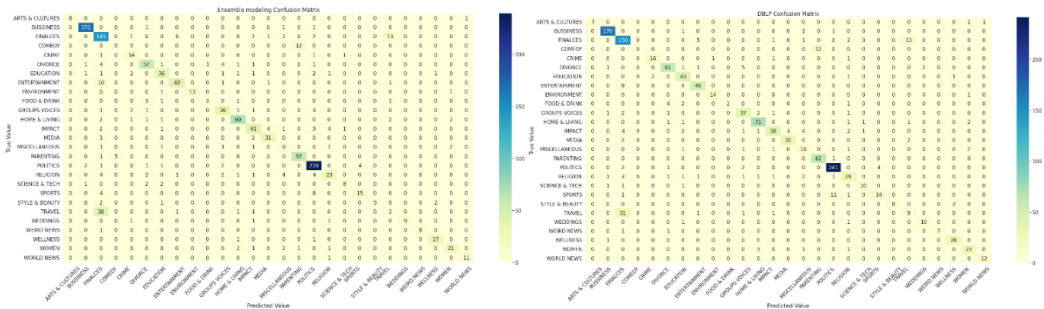

Figure 5 Confusion matrix comparison

### 4.2.2 Comparison with multimodel methods

We conducted comparative experiments on the COVID-Q dataset. The composition of the data is shown in **Figure 6**. where text is the text content of the question, and category represents the category of the question. We used a total of 4113 data points for training in 15 categories, as shown in **Figure 7**. The data for each category are shown in the figure below. It can be seen that the most data have 280 data points, and the least data point has 230 data points. Our data are imbalanced. Regarding the test data, we used a total of 668 pieces of data, which is also unbalanced.

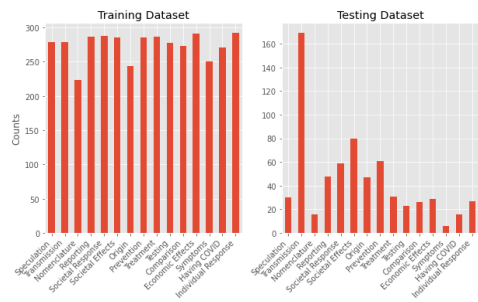

Figure 6 COVID-Q data composition

Figure 7 The distribution of COVID-Q data

From **Table 1**, we can see that the precision of the DBLP model is 60% in 84 minutes, while the best accuracy is 61% based on the Bert + BI-LSTM model, but its training time is also the longest, and it takes 380 minutes. Moreover, we also see that the DISTILBERT-CNN method achieves the fastest training speed, but at the same time, the accuracy is also reduced more. Our method obtains the highest F1 value. It shows that the bidirectional-layer LSTM network can better capture the fine-grained classification of semantic information between phrases and clauses in a small corpus. When DBLP is only 1% lower than the BERT-based method, the time is 100 minutes lower and only 84 minutes.

| Model | Precision | Recall | F1-score | Time | Total params | Trainable Params | Epoch | Size |
|---|---|---|---|---|---|---|---|---|
| BERT+BI-GRU | 60% | 54% | 54% | 241 min | 109,920,145 | 1,609,873 | 6 | 419MB |
| BERT-BI-LSTM | **61%** | 53% | 54% | 380 min | 109,247,003 | 936,731 | 11 | 417MB |
| BERT +CNN | 58% | 51% | 52% | 153 min | 108,404,591 | 94,319 | 4 | 413MB |
| BERT-CNN-LSTM | 57% | 51% | 52% | 301 min | 109,200,743 | 890,471 | 8 | 416MB |
| BERT-LSTM-CNN | 52% | 44% | 44% | 189 min | 111,918,864 | 3,608,592 | 5 | 427MB |
| DISTILBERT + BI-GRU | 59% | 54% | 54% | 85 min | 66,800,785 | 1,609,873 | 5 | 261MB |
| DISTILBERT BI-LSTM | **60%** | **54%** | **54%** | **84 min** | **66,127,643** | **936,731** | **5** | **252MB** |
| DISTILBERT-CNN | 56% | 53% | 53% | **69 min** | 65,285,231 | 94,319 | 4 | 249MB |
| DISTILBERT-CNN-LSTM | 59% | 50% | 51% | 132 min | 66,081,383 | 890,471 | 7 | 252MB |
| DISTILBERT-LSTM-CNN | 56% | 46% | 48% | 75 min | 68,799,504 | 3,608,592 | 4 | 268MB |

**Table 1 Method comparison**

The total parameters of the model with Distilbert decreased significantly, but the trainable parameters were only related to the feature extraction network. The model with the CNN network only achieved the smallest number of trainable params 94319, but its accuracy was also the lowest. This is because CNN cannot well express the information of context and the problem of polysemy representation. On the number of training rounds, we set the early stopping function, which looks at the validation of the model loss score: if there is no improvement after 3 epochs, stop the training. Our model finally stops after 5 epochs. For the final model, because Distilbert has only six layers of encoders, it is approximately half as small as the pretraining model based on BERT.

In summary, the advantages of our model are reflected in the following aspects. It requires fewer trainable parameters and saves computing resources. It can also be easily used by general in-depth learning enthusiasts and is conducive to deployment on the mobile terminal.

### 4.2.3 Performance of real-world data sets

Finally, we collected 4,896 short comments on the network to see how our model performed on the real-world task, with 27 categories in total, and the last category has only 49 data points. We annotated the data using a regularized and artificial approach. The final data have only two dimensions, text and category. We divide the data set into 64% train, 16% val, and 20% test. Thus, we use ranking metrics, such as accuracy and loss, to check our classifier. We compare the ranks produced by our classifier (ensemble classifier and DBLP classifier) to the SVM model, which is suitable for an extremely small dataset. The results are shown in **Figure 8** and **Table 2**.

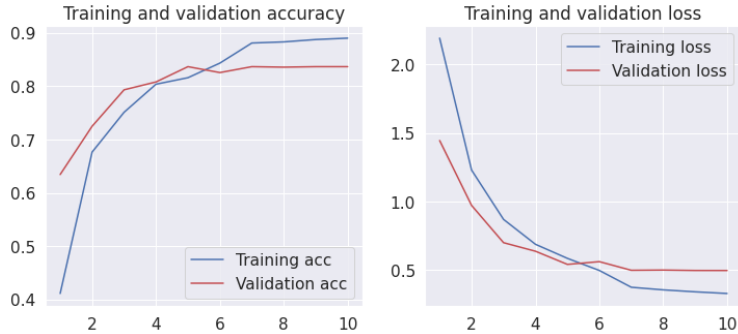

**Figure 8 Training and validation accuracy in real-world data**

| Classification task | Model | Accuracy |
|---|---|---|
| Log Case Classification (27 categories) | SVM | 80.28% |
| | Ensemble model | 84.04% |
| | DBLP | 86.65% |

**Table 2 Comparison of models for log case classification**

The result shows that, for log-case classification, an SVM classifier can solve the problem at some stage. However, SVM is limited to the recognition of the presence of words in the training dataset and is unable to capture the inner relationship of the sentence. In some sentences, the appearance of multiple keywords fooled the SVM model to label this as an 's390' statement, for example, while the negation in the first place completely converted the case. The DBLP can capture the semantic features of the text better than the SVM model and Ensemble model. Overall, the log-case classification was good, with an accuracy of 86.65%. By looking at some mislabelled data, we must admit that these texts are ambiguous and hard to classify.

# 5. Conclusion

The results show that the DBLP model is guaranteed to be able to solve end-to-end data imbalance, small text, and multiclassification tasks and can be directly applied to real-world tasks. Furthermore, we can see that since we utilize the pre-trained model after distillation, the training parameters of our model drop significantly, which is very friendly to general developers. Due to the smaller model size, our model can better apply AI technology to mobile smart devices, such as new energy vehicles and smart home appliances. In contrast, although the BERT-based pretraining model can achieve better accuracy, the computational cost and application cost are high, and it is more suitable for large-scale cloud computing and scientific research scenarios such as research institutes.

In future research, we will improve the model in the following two parts. We combine semantic generation models to complement the important context that may be missing in sentences and apply our model to various tasks on smart devices.

# 6. Reference

1. Kowsari K, Jafari Meimandi K, Heidarysafa M, et al. Text classification algorithms: A survey[J]. Information, 2019, 10(4): 150.
2. Li X. Chinese Language and Literature Online Resource Classification Algorithm Based on Improved SVM[J]. Scientific Programming, 2022, 2022.
3. Vaish K, Deepak G, Santhanavijayan A. DSEORA: Integration of Deep Learning and Metaheuristics for Web Page Recommendation Based on Search Engine Optimization Ranking Emerging Research in Computing, Information, Communication, and Applications. Springer, Singapore, 2022: 873-883.
4. Chen Y Liu D, Liu Y, et al. Research on user-generated content in Q&A system and online comments based on text mining [J]. Alexandria Engineering Journal, 2022
5. Lou H, Zhao H, Deng W. Research on Civil Aviation Passenger Question Intention Recognition Based on Text Classification Method of Self-attention and Deep Neural Network[C] Proceedings of 2021 Chinese Intelligent Automation Conference. Springer, Singapore, 2022: 286-293.
6. Chen H, Wu L, Chen J et al. A comparative study of automated legal text classification using random forests and deep learning [J]. Information Processing & Management, 2022, 59
7. Tarekegn A N, Giacobini M, Michalak K. A review of methods for imbalanced multi-label classification[J]. Pattern Recognition, 2021, 118: 107965.
8. Devlin J, Chang M W, Lee K, et al. Bert: Pre-training of deep bidirectional transformers for language understanding [J]. arXiv preprint arXiv:1810.04805, 2018.
9. Song Y, Wang J, Liang Z, et al. Utilizing BERT intermediate layers for aspect based sentiment analysis and natural language inference[J]. arXiv preprint arXiv:2002.04815, 2020.
10. Wan C X, Li B. Financial causal sentence recognition based on BERT-CNN text classification [J]. The Journal of Supercomputing, 2022, 78(5): 6503-6527.
11. Sanh V, Debut L, Chaumond J, et al. DistilBERT, a distilled version of BERT: smaller, faster, cheaper and lighter[J]. arXiv preprint arXiv:1910.01108, 2019.
12. Xiong G, Yan K. Multitask sentiment classification model based on DistilBert and multiscale CNN [C]//2021 IEEE Intl Conf on Dependable, Autonomic and Secure Computing, Intl Conf on Pervasive Intelligence and Computing, Intl Conf on Cloud and Big Data Computing, Intl Conf on Cyber Science and Technology Congress (DASC/PiCom/CBDCom/CyberSciTech). IEEE, 2021: 700-707.
13. Adel H, Dahou A, Mabrouk A, et al. Improving Crisis Events Detection Using DistilBERT with the Hunger Games Search Algorithm [J]. Mathematics, 2022, 10(3): 447.
14. Ghiassi M, Lee S, Gaikwad S R. Sentiment Analysis and Spam Filtering using the YAC2 Clustering Algorithm with Transferability[J]. Computers & Industrial Engineering, 2022: 107959.
15. Kim Y, Bang S, Sohn J, et al. Question answering method for infrastructure damage information retrieval from textual data using bidirectional encoder representations from transformers[J]. Automation in Construction, 2022, 134: 104061.
16. STITINI O, KALOUN S, BENCHAREF O. Integrating contextual information into multiclass classification to improve the context-aware recommendation [J]. Procedia Computer Science, 2022, 198: 311-316.

17. Yang DU, Kim B, Lee S H, et al. AutoDefect: Defect text classification in residential buildings using a multi-task channel attention network[J]. Sustainable Cities and Society, 2022: 103803.
18. Ma Y, Liu X, Zhao L, et al. Hybrid embedding-based text representation for hierarchical multi-label text classification[J]. Expert Systems with Applications, 2022, 187: 115905.
19. Bilal S F, Almazroi A A, Bashir S, et al. An ensemble based approach using a combination of clustering and classification algorithms to enhance customer churn prediction in telecom industry[J]. PeerJ Computer Science, 2022, 8: e854.
20. Wang K, Liu Y, Cao B et al. TkTC: A framework for top-k text classification of multimedia computing in wireless networks[J]. Wireless Networks, 2022: 1-12.
21. Zhu Y, Zhou X, Qiang J, et al. Prompt-learning for Short Text Classification[J]. arXiv preprint arXiv:2202.11345, 2022.
22. Andersen J S, Maalej W. Efficient, Uncertainty-based Moderation of Neural Networks Text Classifiers[J]. arXiv preprint arXiv:2204.01334, 2022.
23. Chang J W, Yen N, Hung JC. Design of a NLP-empowered finance fraud awareness model: the anti-fraud chatbot for fraud detection and fraud classification as an instance[J]. Journal of Ambient Intelligence and Humanized Computing, 2022: 1-17.
24. Sagi O, Rokach L. Ensemble learning: A survey[J]. Wiley Interdisciplinary Reviews: Data Mining and Knowledge Discovery, 2018, 8(4): e1249.
25. Wei J, Huang C, Vosoughi S, et al. What are people asking about covid-19? a question classification dataset[J]. arXiv preprint arXiv:2005.12522, 2020.
26. Qasim R, Bangyal W H, Alqarni M A, et al. A Fine-Tuned BERT-Based Transfer Learning Approach for Text Classification[J]. Journal of healthcare engineering, 2022, 2022.